\documentclass[runningheads]{llncs}
\usepackage{graphicx}
\usepackage[usenames,dvipsnames]{xcolor}
\usepackage{csquotes}
\usepackage{algorithmic}
\usepackage{csquotes}
\usepackage{cite}
\usepackage{amsmath,amssymb,amsfonts}
\usepackage{algorithmic}
\usepackage{graphicx}
\usepackage{textcomp}

\usepackage{bbold}
\usepackage{caption}

\usepackage{float}
\usepackage{graphicx}
\usepackage{caption}
\usepackage{array}
 \usepackage{subcaption}

\begin{document}
%
\title{A Novel Approach to Train Diverse Types of Language Models for Health Mention Classification of Tweets}
\titlerunning{A Novel Approach to Train Language Models for HMC}
%
\author{Pervaiz Iqbal Khan\inst{1,2}\orcidID{0000-0002-1805-335X} \and
Imran Razzak\inst{3}\orcidID{0000-0002-3930-6600}
\and
Andreas Dengel\inst{1,2}\orcidID{0000-0002-6100-8255} \and
Sheraz Ahmed\inst{1}\orcidID{0000-0002-4239-6520}}
\authorrunning{Pervaiz Iqbal Khan et al.}
%
\institute{German Research Center for Artificial Intelligence (DFKI), Kaiserslautern, Germany \and
TU Kaiserslautern, Kaiserslautern, Germany \and
UNSW, Sydney, Austrailia
\email{\{pervaiz.khan,andreas.dengel,sheraz.ahmed\}@dfki.de}\
\email{imran.razzak@unsw.edu.au}
\
}
\maketitle              
\begin{abstract}
Health mention classification deals with the disease detection in a given text containing disease words. However, non-health and figurative use of disease words adds challenges to the task. Recently, adversarial training acting as a means of regularization has gained popularity in many NLP tasks. In this paper, we propose a novel approach to train language models for health mention classification of tweets that involves adversarial training. We generate adversarial examples by adding perturbation to the representations of transformer models for tweet examples at various levels using Gaussian noise. Further, we employ contrastive loss as an additional objective function. We evaluate the proposed method on the PHM2017 dataset extended version. Results show that our proposed approach improves the performance of classifier significantly over the baseline methods. Moreover, our analysis shows that adding noise at earlier layers improves models' performance whereas adding noise at intermediate layers deteriorates models' performance. Finally, adding noise towards the final layers performs better than the middle layers noise addition.

\keywords{Health informatics  \and Adversarial training \and Health mention classification.}
\end{abstract}

\section{Introduction}
Health Mention Classification (HMC) deals with the detection of disease in a given piece of input text. Authorities can use such classification results to monitor the spread of diseases. Further, early detection of health conditions can help in taking preventive measures and efficiently managing resources in emergencies. To train a classifier for HMC, health-related data is collected from social media platforms such as Twitter, Facebook, etc, based on keywords. These keywords contain the names of the diseases such as fever, heart attack, cancer, etc. However, a keyword-based search may result in noisy data, i.e., collected data may contain disease words but as non-health or figurative mentions. For example, \enquote{Wow, this is awesome it's like having a depression} and \enquote{I nearly had a stroke readin this} are examples of figurative mentions of disease words \emph{depression} and \emph{stroke}. Another tweet \enquote{hearing people cough makes me angry} contains the word \emph{cough} as a non-health mention. Another tweet has a health mention of the disease word  \emph{cancer}: \enquote{my grandpa just diagnosed with lymphoma cancer today, I am devastated and don't know what to do}. These figurative and non-health mentions of disease words make the HMC task challenging. Learning the context of words based on their surrounding words is the key to improving classification results. 

Transformer models\cite{vaswani2017attention} have revolutionized many natural language processing (NLP) tasks\cite{sun2019fine}. The original transformer model consists of encoder-decoder blocks with self-attention layers. Both encoder and decoder blocks consist of multiple encoder and decoder layers, where a specific encoder layer attends to all the layers before and after it to learn word representations during training. On the other hand, decoder layers only attend to the layers before them. BERT\cite{devlin2018bert} is a transformer model that utilizes the encoder block of the original transformer model and is pretrained on the large unlabelled corpus of text. BERT learns the representations of the words by using two different objective functions. 1) Masked Language Modeling (MLM), where random tokens are masked and BERT tries to predict those masked tokens. 2) Next sentence prediction (NSP), where two sentences are given as inputs to the BERT, and it learns to predict whether one sentence follows the other sentence or not. The pretrained model then can be finetuned on downstream tasks such as text classification, etc. RoBERTa\cite{liu2019roberta} is another transformer-based model that has an architecture like BERT, however, unlike BERT, it uses dynamic masking of tokens instead of static tokens. Moreover, it is pretrained on 1000\% more data than BERT.

Adversarial training (AT)\cite{goodfellow2014explaining} is used in many tasks as a regularization technique to improve models' robustness against adversarial attacks\cite{pan2021improved,miyato2016adversarial}. During AT a small perturbation is added to the original input sample and then the model is trained in parallel with both the original and perturbed sample. \cite{miyato2016adversarial} proposed a technique to add perturbation in the word embeddings for the text classification. Recently, self-supervised methods\cite{zbontar2021barlow,chen2020simple,he2020momentum} have gained popularity among researchers in the image processing domain. These methods add perturbations to the inputs, and the training objective is to learn similar representations for the pair of clean and perturbed examples while learning different representations for other examples. Barlow Twins\cite{zbontar2021barlow} is one such method that works on the principle of redundancy reduction. In this paper, we propose a novel approach of training language models for HMC task on Twitter data that combines the ideas of adversarial training and self-supervised learning. Specifically, we add Gaussian noise as a perturbation to the representations of two language models BERT\textsubscript{Large} and RoBERTa\textsubscript{Large} and employ Barlow Twins as an additional loss for learning the similar representations for a pair of clean and perturbed examples. Moreover, instead of perturbing word embeddings, we experiment with adding noise at various layers level. Experiments show that our propose approach improves classification results on both BERT\textsubscript{Large} and RoBERTa\textsubscript{Large} models over their baselines. Further, our analysis shows that adding noise to earlier layers improves models' performance compared to baseline methods, whereas adding noise in the intermediate layers degrades model performance. Finally, adding noise towards the last layers again starts improving performance as compared to intermediate layers. The contributions of this paper towards the adversarial training on diverse transformers models for HMC of tweets are manifold. First, it proposes a new training method for the HMC of tweets using adversarial and contrastive learning methods by adding perturbation to hidden representations. Second, it explores the impacts of noise addition on initial layers of transformer models, then on intermediate layers, and finally on the last layers. Third, it analyzes the impact of noise amount on adversarial training. Fourth, it leverages explainable AI to understand the importance of words in a Tweet for the classification decision.

The rest of the paper is organized as follows: In section \ref{ls}, we discuss the related work, whereas, in section \ref{method}, we present our method for HMC. In section \ref{exp}, we give experimentation detail. In section \ref{raa}, we present results and analysis of the experiments. Finally, in section \ref{conc}, we provide the conclusion of the paper.
\section{Related Work}\label{ls}
\subsection{Adversarial Training}
Adversarial training (AT) has shown success in many computer vision tasks \cite{chen2018robust,song2018physical,xie2017adversarial, xie2017adversarial,arnab2018robustness,goodfellow2014explaining,papernot2016limitations,su2019one}. It is used to increase the model's robustness and safeguard it against the \enquote{malicious} attacks. The process of AT involves training a model simultaneously with a pair of clean and adversarial examples. To generate adversarial examples, perturbations are added using methods such as Gaussian noise and Fast Gradient Sign Method (FGSM)\cite{goodfellow2014explaining}. \cite{miyato2016adversarial} perturbed word embeddings instead of original input text using FGSM for NLP task. Some of the recent works\cite{kitada2021attention,kitada2021making,zhu2019freelb} added perturbations to the attention mechanism of transformer methods using FGSM.\cite{madry2017towards} used multi-step FGSM to generate adversarial examples that proved more effective at the cost of computational overhead. \cite{shafahi2019adversarial} proposed a fast method for adversarial training where perturbations and gradients with respect to parameters of the model were calculated and updated in the same backward pass. This method helped in reducing the cost associated with adversarial training. \cite{zhu2019freelb} proposed \enquote{Free Large-Batch} algorithm in the domain of natural language understanding where perturbations were added in the embedding matrix. Authors observed in-variance in the embedding space that is correlated to the generalization of the method. \cite{miyato2016adversarial} applied perturbations to the word embeddings of recurrent neural network (RNN) embeddings instead of embedding matrix. 
\subsection{Self-Supervised Representation Learning}
Deep learning methods automatically learn a mapping between input and output samples when given sufficient training samples. AlexNet\cite{krizhevsky2012imagenet} was the first deep learning model trained end-to-end on ImageNet\cite{russakovsky2015imagenet} dataset for image classification and reduced the classification error significantly as compared to previous methods.

The success for supervised deep learning methods depends on the availability of largely annotated data that is costly in practice. Self-supervised learning (SSL) methods have gained popularity where these methods learn representations from unlabeled data. In NLP, many language models \cite{yang2019xlnet,devlin2018bert,liu2019roberta} learned representations from the large unlabelled corpus of text\cite{zhu2015aligning, parker2011english, callan2012lemur, crawl2019common} by defining proxy tasks such as masked language prediction. Similarly, SSL methods has shown success in computer vision. Momentum Contrast (MoCo)\cite{he2020momentum} method worked on the idea of moving-average encoder. To learn feature representations, SimCLR\cite{chen2020simple} first drastically augmented the images and then trained the model to maximize the cosine similarity between original images and their augmented versions while pushing the other images away from them. SimCLR used in-batch negative samples and heavily relied on large batch sizes. Bootstrap Your Own Latent (BYOL)\cite{grill2020bootstrap} used two versions of the same network called online and target network to learn visual representations. Parameters of the target network were moving average of the online network, therefore, it did not learn new parameters. Each network utilized a different augmented version of the original image. The online network aimed to learn representations similar to the target network representations. BYOL worked well on smaller batches as it did not depend on negative samples. Barlow Twins \cite{zbontar2021barlow} aimed at the principle of redundancy reduction to learn noise-invariant representations. Barlow Twins also worked well on the smaller batches.   
\subsection{Health Mention Classification of Tweets}
\cite{karisani2018did} presented a method called \enquote{WESPAD} acting as a regularizer for HMC task on Twitter data. It partitioned and distorted embedding which helped model in achieving generalization capability. \cite{jiang2018identifying} used non-contextual embeddings for representing the tweets. These embeddings were passed to LSTMs\cite{hochreiter1997long}. Authors showed that the use of LSTM before the classification layer improved the performance compared to the simple SVM, KNN, and Decision Trees. \cite{iyer2019figurative} incorporated features from an unsupervised statistical learner for idiom detection, and passed it to CNN based classifier. The incorporation of these features improved the classification performance over the CNN classifier trained on pre-trained embeddings. \cite{biddle2020leveraging} experimented with both non-contextual embeddings such as word2vec\cite{mikolov2013distributed} and contextual embeddings such as ELMO\cite{peters2018deep} and BERT and incorporated sentiment information using WordNet\cite{baccianella2010sentiwordnet}, VAD\cite{mohammad2018obtaining}, and ULMFit\cite{howard2018universal} for HMC of tweets. \cite{khan2020improving} applied permutation-based pretrained embeddings and finetuned the pretrained model\cite{yang2019xlnet} on the HMC dataset that improved the performance over the existing methods. \cite{khan2022performance} compared the performance of various transformer models on HMC of tweets and showed that RoBERTa\textsubscript{Large} outperformed other methods. 

In this work, we propose a new training approach of language models for the HMC of Tweets that combines the concepts of adversarial training and self-supervised methods. To generate adversarial examples, we add Gaussian noise instead of using FGSM. Instead of adding perturbations word embeddings matrix, we add noise to the hidden representations of two transformer-based models i.e., BERT\textsubscript{Large} and RoBERTa\textsubscript{Large}. Further, we study the impact of adding noise at various layers as well as using Barlow Twins as an additional contrastive loss and analyze how it works compared to the baseline methods.
\section{Methodology}\label{method}
In this section, first, we discuss adversarial training for health mention classification. Then, we discuss the contrastive loss, i.e., Barlow Twins. Finally, we present how we combine adversarial training with the Barlow Twins loss for the health mention classification of tweets.
\subsection{Adversarial Training}\label{at}
Let $x$ be the input example and $L \in \{1, 4, 7, 10, 13, 16, 19, 22\}$ represents the intermediate layer numbers of the transformer model. Let $\eta$ denotes a Gaussian noise with mean `$\mu$' and variance `$\sigma$' given as follows:\\
\begin{equation}
    \eta = N(\mu,\sigma)
\end{equation}
\\

then, we generate adversarial example $x_{adv}$ by adding $\eta$ to the representations of one of the layers in $L$ as given below:\\
\begin{equation}
    x_{adv}= E\textsubscript{L\textsubscript{i}} + \eta
\end{equation}
\\
where $L_i$ denotes $i^{th}$ layer from $L$, and a $E\textsubscript{L\textsubscript{i}}$ represents the embedding of $L_i$. 

We train the model simultaneously on $x$, and $x_{adv}$, and calculate two cross-entropy losses separately on $x$, and $x_{adv}$. 

\subsection{Barlow Twins Loss}
Barlow Twins loss presented in \cite{zbontar2021barlow} jointly operates on two embeddings, one from the original input and other from distorted input. It is based on redundancy reduction principle. Let $E^{clean}$, and $E^{adv}$ represent the embeddings of clean examples and adversarial examples, respectively. Then, $E^{clean}$, and $E^{adv}$ are fed into neural network $f_\theta$, where $\theta$ is a trainable parameter. The outputs of the $f_\theta$ for $E^{clean}$, and $E^{adv}$ are their projections to lower dimensions and centered with mean $0$ across batch dimension. Barlow Twins is defined as given below \cite{zbontar2021barlow}:\\
\begin{equation}
 \mathcal{L} = \sum_{i=1} (1 - M_{ii})^2 + \lambda \sum_{i=1} \sum_{j \neq i}M_{ij}^2
\end{equation}
where $\mathcal{L}$ is a Barlow Twins, $\sum_{i=1} (1 - M_{ii})^2$, and $\sum_{i=1} \sum_{j \neq i}M_{ij}^2$ are invariance, and redundancy reduction terms respectively, and $\lambda$ controls the weight of the two terms. $M$ is a square matrix and computes the cross-correlation between $E^{clean}$, and $E^{adv}$. Values of $M$ vary between -1 (that represents perfect anti-correlation), and +1 (represents perfect correlation). $M_{ij}$ is computed as follows:
\\
\begin{equation}
    M_{ij} = \frac{\sum_{b=1}^{N} E^{clean}_{ b, i} E_{b, i}^{adv} }{\sqrt{ \sum_{b=1}^{N} (E^{clean}_{b, i} })^2    \sqrt{ \sum_{b=1}^{N} (E^{adv}_{b, i} })^2 }
\end{equation}

where $i,j$, represents the index of the matrix $M$, and $b$ represents batch samples.
\subsection{Adversarial Training with Barlow Twins for HMC}
\begin{figure*}[!t] \centering{\includegraphics[width=10cm ,keepaspectratio]{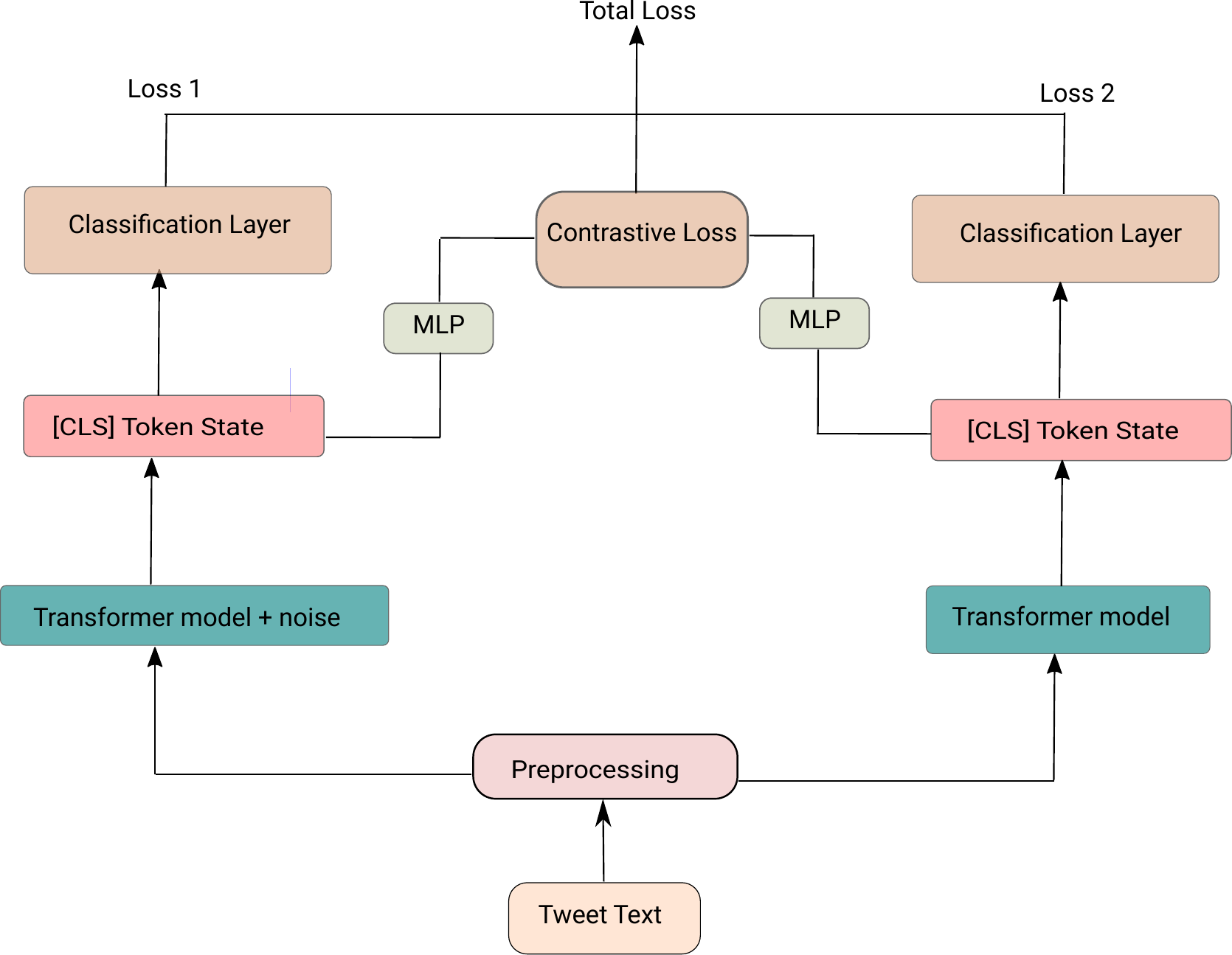}} \caption{Architecture diagram of our proposed training approach.} \label{method-diagram} \end{figure*}

Fig. \ref{method-diagram} shows the architecture diagram of the proposed method. First, we pass input tweet text through a preprocessing step that removes URLs, user mentions, hashtags, and special characters, and converts emojis to their corresponding text representation. After that, we give the preprocessed input to the two transformer models of the same type. The first model directly processes the input example called clean example, whereas the second model adds Gaussian noise $\eta$ with $\mu = 0$ and $\sigma = 1$ to one of the hidden states in layer $L$ to generate an adversarial example (discussed in \ref{at}). Then, we take the embedding of $[CLS]$ token to extract sentence embedding separately for clean and adversarial input example and pass it to the classification layer. We compute cross-entropy losses for each of the clean and adversarial examples separately. We also employ Barlow Twins loss as a third loss. The inputs to the Barlow Twins loss are $[CLS]$ token representations from the clean and adversarial examples projected to lower dimensions by a neural network of two layers. Then we take the weighted average of two cross-entropy losses and a Barlow Twins loss to train the model as given below:
\begin{equation}
\mathcal{L}_{total} = \frac{(1- C )}{2} (\mathcal{L}_{clean} + \mathcal{L}_{adv}) + C \mathcal{L}_{BT}
\end{equation}
 where $\mathcal{L}_{total}$ represents the total loss, $\mathcal{L}_{clean}$ and $\mathcal{L}_{adv}$ represent cross-entropy losses for the clean and adversarial examples, respectively, and $\mathcal{L}_{BT}$ represents Barlow Twins loss. `C' is the trade-off parameter between three losses.

\section{Experiments}\label{exp}
\subsection{Dataset}\label{DS}
We perform experimentation on the extended version of the PHM2017 dataset used in the paper\cite{biddle2020leveraging} which consists of tweets related to $10$ diseases. The dataset contained $15,742$ tweets at download time, out of which $4,228$ tweets are related to health mentions, $7,322$ tweets are non-health mentions, and $4,192$ tweets are figurative mentions. Alzheimer's has the highest number of total tweets, i.e., $1,715$, whereas headache has the lowest number of total tweets, i.e., $1,429$. For non-health mention tweets, heart attack contains the lowest number of tweets with a count of $209$, whereas Migraine has the highest number of tweets with a total of $904$. For non-health mention of tweets, Parkinson's and headache contain a maximum and a minimum number of tweets with a count of $1,362$ and $112$, respectively. In the case of figurative mentions, Alzheimer's contains the smallest number of tweets with $92$ examples, whereas heart attack has the highest number of tweets with $1,060$ examples. For experiments, we follow the same train/validation/test split as in the paper \cite{khan2022improving}. Further, we combine the figurative and non-health mention tweets in a single class that reduces the task to binary classification.

\subsection{Experimental Setup}
\subsubsection{Baseline}
For experiments on the PHM2017 dataset, we use pretrained BERT\textsubscript{Large}  and RoBERTa\textsubscript{Large} models and finetune them for the classification task. We take the representations of $[CLS]$ and pass them to the feed-forward layer to classify the input tweet.

\subsubsection{Hyperparameters}
For both the BERT\textsubscript{Large} and RoBERTa\textsubscript{Large}, we set the maximum sequence length to 64, and a fixed learning rate of $1e^{-5}$ for all the experiments. We use AdamW\cite{loshchilov2018fixing} as an optimizer. For all the experiments, we use the batch sizes of $16, 24$, and $32$ and choose the best-performing model on the validation set to evaluate on the test set. In the case of adversarial training combined with contrastive loss, we experiment with a $C \in \{0.1, 0.2, 0.3, 0.4\}$ as a trade-off parameter between two cross-entropy and Barlow Twins losses. However, in the case of adversarial training without Barlow Twins, we give equal weights to the two cross-entropy losses. We finetune both the models for $10$ epochs and use early-stopping to prevent overfitting of the models. Unlike the original implementation of Barlow Twins, we project the original embeddings dimensions of $1024$ to a lower-dimensional space of $300$ that is similar to SimCLR, and proved more effective in our experiments. The projection network consists of $3$ linear layers where the input and output dimensions of the first $2$ layers are $1024$, whereas the output layer dimensions are $300$. The first two linear layers follow 1-d batch normalization and ReLU as an activation function. For Barlow Twins loss, we set its default hyperparameters values as in the original paper implementation.

\section{Results and Analysis}\label{raa}
\begin{figure*}[!t] \centering{\includegraphics[width=12cm ,keepaspectratio]{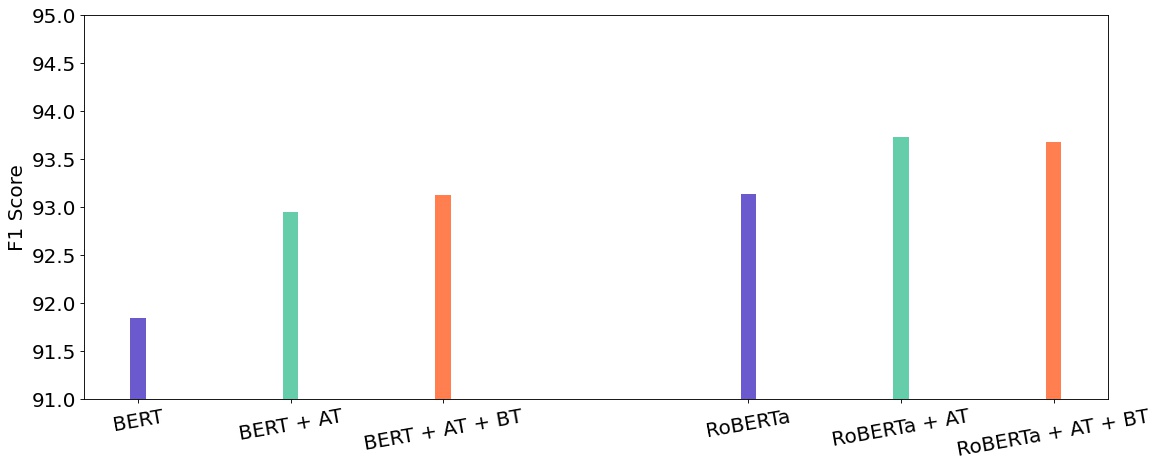}} \caption{Performance of BERT\textsubscript{Large} and RoBERTa\textsubscript{Large} baseline methods, their best results on adversarial training on various layers, and best results on adversarial training combined with Barlow Twins as a contrastive loss.} \label{bar-plot} \end{figure*}

We experiment with the two transformer-based models named BERT\textsubscript{Large} and RoBERTa\textsubscript{Large} to validate our proposed approach. First, we finetune BERT\textsubscript{Large} and RoBERTa\textsubscript{Large} end-to-end for HMC as baseline methods. Then, we perform adversarial training for both models. Further, we additionally use Barlow Twins as a contrastive loss. We compare the results for three settings, i.e., Baseline methods, adversarial training (AT), and adversarial training combined with Barlow Twins loss (AT +BT). Fig \ref{bar-plot} shows the results of all three settings where the BERT\textsubscript{Large} baseline gives an F1 score of 91.84\% whereas BERT\textsubscript{Large} + AT, and BERT\textsubscript{Large} + AT + BT give the best F1 scores of $92.94$\% and $93.12$\%, respectively. In the case of RoBERTa\textsubscript{Large} baseline, we get an F1 score of 93.13\%, whereas RoBERTa\textsubscript{Large} + AT, and  RoBERTa\textsubscript{Large} + AT + BT give the best F1 scores of $93.73$\%, and $93.67$\%, respectively. 

\begingroup 
\setlength{\tabcolsep}{2.0pt} 
\renewcommand{\arraystretch}{1.5}

\subsection{Effect of noise on layers level of models}

Table \ref{results-detail} shows layer-wise results for the experimental settings of adversarial training, and adversarial training with Barlow Twins as contrastive loss. For BERT\textsubscript{Large} + AT, adding noise at the 1st layer gives an F1 score of 92.94\%. Adding noise after the first layer decreases the performance of the model till the 10th layer. After that starting from layer 13th, Gaussian noise gradually starts increasing the model's performance until we reach the 22nd layer. In the case of BERT\textsubscript{Large} + AT + BT, adding noise at layer no. 4 gives the highest F1 score of 93.12\%, then performance starts decreasing till layer no. 16. However, after layer no. 16, the F1 score again starts increasing. For RoBERTa\textsubscript{Large} + AT adding noise at layer no. 1 improves the F1 score to 93.64\%. After that, model performance varies between layer no. 4 and layer no. 10. Then, at layer 13th, the model gives the highest F1 score of 93.73\%. At layer 16th F1 score slightly decreases, and then at layer 19th, layer 22nd  F1 score slightly increases. In the case of RoBERTa\textsubscript{Large} + AT + BT, layer no. 1 gives an F1 score of 93.67\% after that model's performance decreases till layer no. 10. Then starting from layer no. 13, the model's performance increases as compared to layer no. 7 and layer no. 10. The trend we see in the results is that adding noise to earlier layers increases the model's performance most of the time, then in the middle layers, it degrades the model's performance, then finally model's performance starts rising towards the last layers. The reason for the better performance of noise at earlier layers is, that the model has enough layers to recover the hidden representations, and adding noise at these layers increases the model's generalization capability. Similarly, adding noise at the last layers works better than the middle layers because the model has already learned useful representations, and adding small perturbations doesn't harm the model's performance much. However, adding noise at intermediate layers somehow deteriorates the model's performance because the model hasn't learned useful representations, and adding noise at this stage doesn't allow the model to recover from the damage caused at this stage.  BERT\textsubscript{Large} + AT + BT improves the model's performance as compared to the BERT\textsubscript{Large} + AT, however, RoBERTa\textsubscript{Large} + AT + BT performance is sometime slightly worse than the RoBERTa\textsubscript{Large} + AT during intermediate layers and it is better than RoBERTa\textsubscript{Large} + AT in earlier and last layers.
\begin{table}
\begin{center}
\caption{F1 score of AT, and AT + BT for BERT\textsubscript{Large} and RoBERTa\textsubscript{Large} while adding noise at different layers. Results show that most of the time, AT, and AT + BT at earlier layers perform better than the intermediate and last layers.}\label{results-detail}
\begin{tabular}{lllllllll}
\hline
Model &   L\#1 & L\#4 & L\#7 & L\#10 & L\#13 & L\#16 & L\#19 & L\#22 \\
\hline
BERT\textsubscript{Large} + AT &  \textbf{92.94} & 92.66 & 92.12 & 92.03 & 92.41 & 92.43 & 92.40 & 92.74 \\
\hline
BERT\textsubscript{Large} + AT+ BT  & 92.64 & \textbf{93.12} & 92.55 & 92.73 & 92.16 & 91.88 & 92.40 & 93.03 \\
\hline
RoBERTa\textsubscript{Large} + AT &  93.64 & 93.16 & 93.41 & 92.70 & \textbf{93.73} & 93.41 & 93.46 & 93.46 \\
\hline
RoBERTa\textsubscript{Large} + AT + BT & \textbf{93.67} & 93.54 & 92.88 & 92.43 & 93.38 & 93.62 & 93.17 & 93.47 \\
\hline 
\end{tabular} 
\end{center}
\end{table}
\endgroup

\subsection{Effect of noise amount on model's performance}

\begin{figure*}{}

\centering
\begin{subfigure}{0.50\textwidth}
  \centering
  \includegraphics[width=0.99\linewidth]{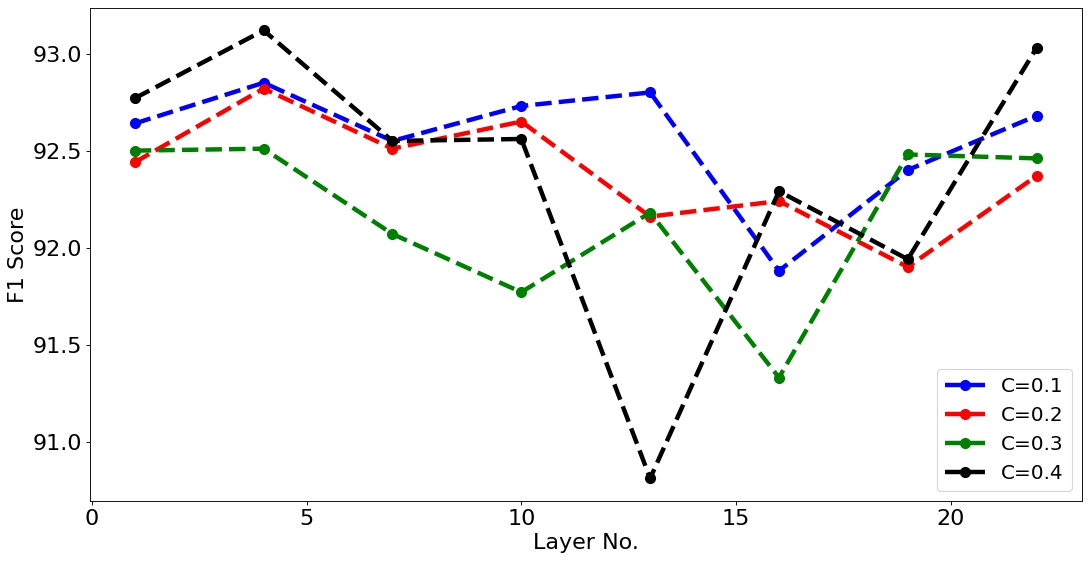}
  \caption{Effect of `C' on BERT\textsubscript{Large} \\ + AT + BT}
\label{fig:noise-amount-bert}
\end{subfigure}%
\begin{subfigure}{0.50\textwidth}
  \centering
  \includegraphics[width=0.99\linewidth]{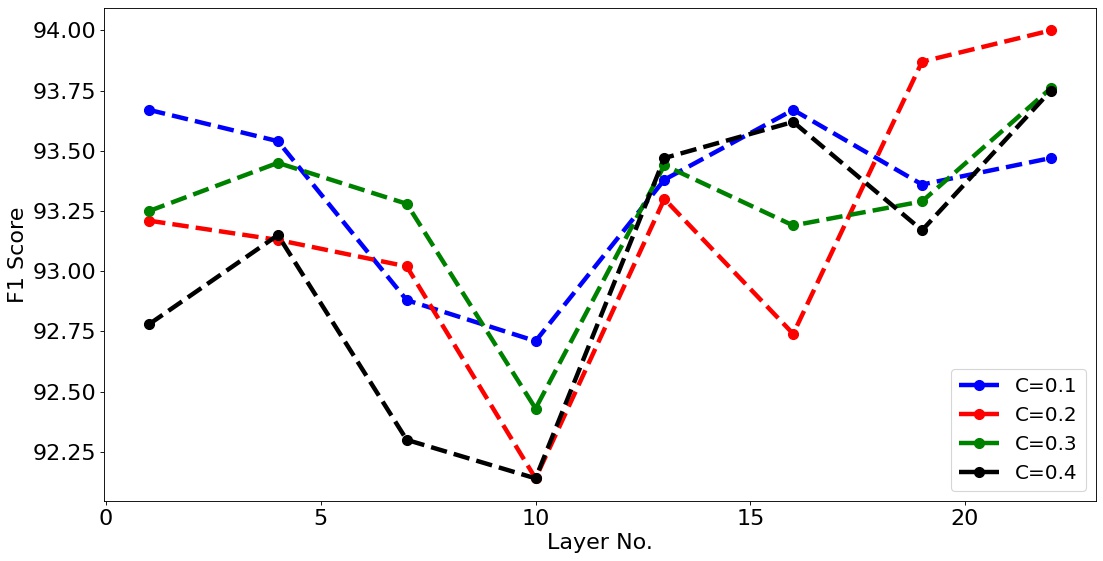}
  \caption{Effect of `C' on RoBERTa\textsubscript{Large} + AT + BT}
  \label{fig:noise-amount-roberta}

\end{subfigure}
\caption{Effect of noise parameter `C' on the performance of both BERT\textsubscript{Large} and RoBERTa\textsubscript{Large} with adversarial training and Barlow Twins loss on test set of PHM2017 dataset.}
  \label{fig:noise-amount-both}
\end{figure*}

Fig. \ref{fig:noise-amount-both} visualizes the effect of noise parameter `C' on models performance. Fig. \ref{fig:noise-amount-bert} shows that at $C=0.4$, layer no. 1 and 4 perform better than other values of `C', and then at layer no. 22, it gives the best F1 score of $93.03$ \% for BERT\textsubscript{Large} + AT + BT. At Layer no. 12, $C=0.4$ gives the lowest F1 score as compared to other values of `C' and other layers. $C=0.1$ performs better than its other values in the middle layers. On $C=0.2$, and $C=0.3$ classifier do not perform well as compared to other values.
Fig. \ref{fig:noise-amount-roberta} shows the effect of noise amount on RoBERTa\textsubscript{Large} + AT + BT performance. $C=0.1$ works well on the initial layers, whereas $C=0.2$ does not perform well on the initial layers, however, it outperforms other values of `C' towards layer no. 19, and 22. $C=0.4$ does not perform well most of the times for RoBERTa\textsubscript{Large} + AT + BT as compared to other values of `C'.

\subsection{Comparison with SOTA}
To compare with state-of-the-art (SOTA), we select the best hyperparameters, i.e. layer no. value of `C', and batch size from the validation set of experiments given in Table \ref{results-detail}, and train models 10-fold cross-validation. As shown in Table \ref{tbl:sota}, our propose method for RoBERTa\textsubscript{Large} + AT + BT beats the state-of-the-art methods in terms of precision, and F1 score. 
\begingroup
\setlength{\tabcolsep}{2pt} 
\renewcommand{\arraystretch}{1.3}
 \begin{table}[!htbp]
\begin{center}
\caption{Comparison of our method with SOTA. Results are the average of Precision (P), Recall (R), and F1 score (F1) on the 10-fold cross-validation of PHM2017 dataset. Our method is directly comparable to methods in \cite{khan2020improving} and \cite{khan2022improving} as dataset distribution does not match with other methods.}
\label{tbl:sota}
\begin{tabular}{llll}
\hline
Method     & P & R & F1\\
\hline
Jiang et al.\cite{jiang2018identifying} & 72.1 & \textbf{95} & 81.8\\
\hline
Karisani et al.\cite{karisani2018did}  & 75.2 & 89.6 & 81.8\\
\hline
Biddle et al.\cite{biddle2020leveraging}  & 75.6 & 92 & 82.9\\
\hline
Khan et al.\cite{khan2020improving}  & 89.1 & 88.2 & 88.4\\
\hline

Khan et al. BERT\textsubscript{Large}\cite{khan2022improving}  & 93.25 & 93.75 & 93.45\\
\hline

Khan et al.  RoBERTa\textsubscript{Large}\cite{khan2022improving} & 93.95 & 94.4 & 94.2\\
\hline

BERT\textsubscript{Large} + AT + BT (ours)  & 93.05 & 93.4 & 93.1\\
\hline

RoBERTa\textsubscript{Large} + AT + BT (ours) & \textbf{94.35} & 94.4 & \textbf{94.45}\\
\hline
\end{tabular}
\end{center}
\end{table}
\endgroup

\subsection{Explaining the model decision}
\begin{table}[!htbp]
\begin{center}
\caption{Visualization of word importance in classifier decision. Words highlighted with green color supported model prediction, and words highlighted with red color opposed model prediction. Classification decision is made by computing the overall score of words in a Tweet. In header columns, GT stands for ground truth, whereas HM and NHM stand for non-health mention and health mention, respectively.}
\renewcommand{\arraystretch}{2.0}
\setlength{\tabcolsep}{2pt} 
\label{tbl:vis}
\begin{tabular}{lll m{6cm}l}
\hline
GT & Prediction & Model            & Word Importance \\
\hline
HM           & NHM              & RoBERTa\textsubscript{Large} baseline  & \includegraphics[width=6cm]{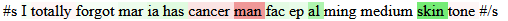}            \\
             & HM               & RoBERTa\textsubscript{Large} + AT + BT      & \includegraphics[width=6cm ]{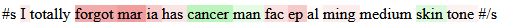}               \\

\hline    
HM           & NHM              & RoBERTa\textsubscript{Large} baseline  & \includegraphics[width=4cm ]{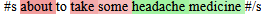}            \\
             & HM               & RoBERTa\textsubscript{Large} + AT + BT      & \includegraphics[width=4cm ]{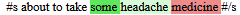}               \\

\hline    
NHM           & HM              & RoBERTa\textsubscript{Large} baseline & \includegraphics[width=6cm ]{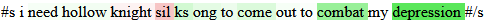}               \\
             & HHM               & RoBERTa\textsubscript{Large} + AT + BT         & \includegraphics[width=6cm ]{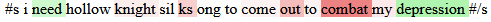}               \\
\hline  
NHM           & HM              & RoBERTa\textsubscript{Large} baseline & \includegraphics[width=6cm ]{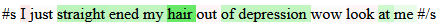}               \\
             & HHM               & RoBERTa\textsubscript{Large} + AT + BT         & \includegraphics[width=6cm ]{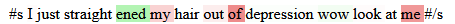}               \\
\hline     
\end{tabular}
\end{center}
\end{table}
To analyze the words that play a significant role in model classification decisions, we utilize a transformer interpret library\cite{kokhlikyan2020captum} that is based on \emph{Layer Integrated Gradients} algorithm\cite{sundararajan2017axiomatic}. For this purpose, we visualize some of the random examples from the test set that are misclassified by the baseline method and correctly classified by our proposed method. Further, we highlight the words supporting the model's decisions as green and words opposing the model's decisions as red. As shown in Table \ref{tbl:vis}, the first tweet in the table is correctly classified by RoBERTa\textsubscript{Large} + AT + BT and misclassified by RoBERTa\textsubscript{Large} baseline model. Similarly,  the second tweet is a health mention and correctly classified by  RoBERTa\textsubscript{Large} + AT + BT and misclassified by RoBERTa\textsubscript{Large} baseline model. The third and fourth tweets are non-health mentions and misclassified by RoBERTa\textsubscript{Large} baseline and correctly classified by RoBERTa\textsubscript{Large} + AT + BT.

\section{Conclusion}\label{conc}
In this paper, we presented a new approach for HMC of tweet examples that combines adversarial training with the contrastive loss. We employed Gaussian noise with mean $0$ and standard deviation of $1$ at various internal representations levels of two transformer models BERT\textsubscript{Large} and RoBERTa\textsubscript{Large} and utilized Barlow Twins as a contrastive loss. We evaluated our method on PHM2017 dataset extended version, and the results showed that our proposed approach improved performance over the baseline methods. Further analysis showed that adding noise at initial layers improved models' performance over baseline, whereas noise addition at intermediate layers decreased models' performance. Finally, we observed that adding noise towards the final layers performed better than the noise at intermediate layers.

%
%

%
%
%
%
\bibliographystyle{splncs04}
\bibliography{references}

\begin{thebibliography}{10}
\providecommand{\url}[1]{\texttt{#1}}
\providecommand{\urlprefix}{URL }
\providecommand{\doi}[1]{https://doi.org/#1}

\bibitem{arnab2018robustness}
Arnab, A., Miksik, O., Torr, P.H.: On the robustness of semantic segmentation
  models to adversarial attacks. In: Proceedings of the IEEE Conference on
  Computer Vision and Pattern Recognition. pp. 888--897 (2018)

\bibitem{baccianella2010sentiwordnet}
Baccianella, S., Esuli, A., Sebastiani, F.: Sentiwordnet 3.0: an enhanced
  lexical resource for sentiment analysis and opinion mining. In: Lrec.
  vol.~10, pp. 2200--2204 (2010)

\bibitem{biddle2020leveraging}
Biddle, R., Joshi, A., Liu, S., Paris, C., Xu, G.: Leveraging sentiment
  distributions to distinguish figurative from literal health reports on
  twitter. In: Proceedings of The Web Conference 2020. pp. 1217--1227 (2020)

\bibitem{callan2012lemur}
Callan, J.: The lemur project and its clueweb12 dataset. In: Invited talk at
  the SIGIR 2012 Workshop on Open-Source Information Retrieval (2012)

\bibitem{chen2018robust}
Chen, S.T., Cornelius, C., Martin, J., Chau, D.H.: Robust physical adversarial
  attack on faster r-cnn object detector. corr abs/1804.05810 (2018). arXiv
  preprint arXiv:1804.05810  (2018)

\bibitem{chen2020simple}
Chen, T., Kornblith, S., Norouzi, M., Hinton, G.: A simple framework for
  contrastive learning of visual representations. In: International conference
  on machine learning. pp. 1597--1607. PMLR (2020)

\bibitem{crawl2019common}
Crawl, C.: Common crawl corpus. Online at http://commoncrawl. org  (2019)

\bibitem{devlin2018bert}
Devlin, J., Chang, M.W., Lee, K., Toutanova, K.: Bert: Pre-training of deep
  bidirectional transformers for language understanding. arXiv preprint
  arXiv:1810.04805  (2018)

\bibitem{goodfellow2014explaining}
Goodfellow, I.J., Shlens, J., Szegedy, C.: Explaining and harnessing
  adversarial examples. arXiv preprint arXiv:1412.6572  (2014)

\bibitem{grill2020bootstrap}
Grill, J.B., Strub, F., Altch{\'e}, F., Tallec, C., Richemond, P., Buchatskaya,
  E., Doersch, C., Avila~Pires, B., Guo, Z., Gheshlaghi~Azar, M., et~al.:
  Bootstrap your own latent-a new approach to self-supervised learning.
  Advances in Neural Information Processing Systems  \textbf{33},  21271--21284
  (2020)

\bibitem{he2020momentum}
He, K., Fan, H., Wu, Y., Xie, S., Girshick, R.: Momentum contrast for
  unsupervised visual representation learning. In: Proceedings of the IEEE/CVF
  conference on computer vision and pattern recognition. pp. 9729--9738 (2020)

\bibitem{hochreiter1997long}
Hochreiter, S., Schmidhuber, J.: Long short-term memory. Neural computation
  \textbf{9}(8),  1735--1780 (1997)

\bibitem{howard2018universal}
Howard, J., Ruder, S.: Universal language model fine-tuning for text
  classification. arXiv preprint arXiv:1801.06146  (2018)

\bibitem{iyer2019figurative}
Iyer, A., Joshi, A., Karimi, S., Sparks, R., Paris, C.: Figurative usage
  detection of symptom words to improve personal health mention detection.
  arXiv preprint arXiv:1906.05466  (2019)

\bibitem{jiang2018identifying}
Jiang, K., Feng, S., Song, Q., Calix, R.A., Gupta, M., Bernard, G.R.:
  Identifying tweets of personal health experience through word embedding and
  lstm neural network. BMC bioinformatics  \textbf{19}(8), ~210 (2018)

\bibitem{karisani2018did}
Karisani, P., Agichtein, E.: Did you really just have a heart attack? towards
  robust detection of personal health mentions in social media. In: Proceedings
  of the 2018 World Wide Web Conference. pp. 137--146 (2018)

\bibitem{khan2020improving}
Khan, P.I., Razzak, I., Dengel, A., Ahmed, S.: Improving personal health
  mention detection on twitter using permutation based word representation
  learning. In: International Conference on Neural Information Processing. pp.
  776--785. Springer (2020)

\bibitem{khan2022performance}
Khan, P.I., Razzak, I., Dengel, A., Ahmed, S.: Performance comparison of
  transformer-based models on twitter health mention classification. IEEE
  Transactions on Computational Social Systems  (2022)

\bibitem{khan2022improving}
Khan, P.I., Siddiqui, S.A., Razzak, I., Dengel, A., Ahmed, S.: Improving health
  mentioning classification of tweets using contrastive adversarial training.
  arXiv preprint arXiv:2203.01895  (2022)

\bibitem{kitada2021attention}
Kitada, S., Iyatomi, H.: Attention meets perturbations: Robust and
  interpretable attention with adversarial training. IEEE Access  \textbf{9},
  92974--92985 (2021)

\bibitem{kitada2021making}
Kitada, S., Iyatomi, H.: Making attention mechanisms more robust and
  interpretable with virtual adversarial training for semi-supervised text
  classification. arXiv preprint arXiv:2104.08763  (2021)

\bibitem{kokhlikyan2020captum}
Kokhlikyan, N., Miglani, V., Martin, M., Wang, E., Alsallakh, B., Reynolds, J.,
  Melnikov, A., Kliushkina, N., Araya, C., Yan, S., et~al.: Captum: A unified
  and generic model interpretability library for pytorch. arXiv preprint
  arXiv:2009.07896  (2020)

\bibitem{krizhevsky2012imagenet}
Krizhevsky, A., Sutskever, I., Hinton, G.E.: Imagenet classification with deep
  convolutional neural networks. Advances in neural information processing
  systems  \textbf{25} (2012)

\bibitem{liu2019roberta}
Liu, Y., Ott, M., Goyal, N., Du, J., Joshi, M., Chen, D., Levy, O., Lewis, M.,
  Zettlemoyer, L., Stoyanov, V.: Roberta: A robustly optimized bert pretraining
  approach. arXiv preprint arXiv:1907.11692  (2019)

\bibitem{loshchilov2018fixing}
Loshchilov, I., Hutter, F.: Fixing weight decay regularization in adam. arXiv
  preprint arXiv:2011.08042v1  (2018)

\bibitem{madry2017towards}
Madry, A., Makelov, A., Schmidt, L., Tsipras, D., Vladu, A.: Towards deep
  learning models resistant to adversarial attacks. arXiv preprint
  arXiv:1706.06083  (2017)

\bibitem{mikolov2013distributed}
Mikolov, T., Sutskever, I., Chen, K., Corrado, G.S., Dean, J.: Distributed
  representations of words and phrases and their compositionality. In: Advances
  in neural information processing systems. pp. 3111--3119 (2013)

\bibitem{miyato2016adversarial}
Miyato, T., Dai, A.M., Goodfellow, I.: Adversarial training methods for
  semi-supervised text classification. arXiv preprint arXiv:1605.07725  (2016)

\bibitem{mohammad2018obtaining}
Mohammad, S.: Obtaining reliable human ratings of valence, arousal, and
  dominance for 20,000 english words. In: Proceedings of the 56th Annual
  Meeting of the Association for Computational Linguistics (Volume 1: Long
  Papers). pp. 174--184 (2018)

\bibitem{pan2021improved}
Pan, L., Hang, C.W., Sil, A., Potdar, S., Yu, M.: Improved text classification
  via contrastive adversarial training. arXiv preprint arXiv:2107.10137  (2021)

\bibitem{papernot2016limitations}
Papernot, N., McDaniel, P., Jha, S., Fredrikson, M., Celik, Z.B., Swami, A.:
  The limitations of deep learning in adversarial settings. In: 2016 IEEE
  European symposium on security and privacy (EuroS\&P). pp. 372--387. IEEE
  (2016)

\bibitem{parker2011english}
Parker, R., Graff, D., Kong, J., Chen, K., Maeda, K.: English gigaword fifth
  edition ldc2011t07 (tech. rep.). Tech. rep., Technical Report. Linguistic
  Data Consortium, Philadelphia (2011)

\bibitem{peters2018deep}
Peters, M.E., Neumann, M., Iyyer, M., Gardner, M., Clark, C., Lee, K.,
  Zettlemoyer, L.: Deep contextualized word representations. arXiv preprint
  arXiv:1802.05365  (2018)

\bibitem{russakovsky2015imagenet}
Russakovsky, O., Deng, J., Su, H., Krause, J., Satheesh, S., Ma, S., Huang, Z.,
  Karpathy, A., Khosla, A., Bernstein, M., et~al.: Imagenet large scale visual
  recognition challenge. International journal of computer vision
  \textbf{115}(3),  211--252 (2015)

\bibitem{shafahi2019adversarial}
Shafahi, A., Najibi, M., Ghiasi, M.A., Xu, Z., Dickerson, J., Studer, C.,
  Davis, L.S., Taylor, G., Goldstein, T.: Adversarial training for free!
  Advances in Neural Information Processing Systems  \textbf{32} (2019)

\bibitem{song2018physical}
Song, D., Eykholt, K., Evtimov, I., Fernandes, E., Li, B., Rahmati, A., Tramer,
  F., Prakash, A., Kohno, T.: Physical adversarial examples for object
  detectors. In: 12th $\{$USENIX$\}$ Workshop on Offensive Technologies
  ($\{$WOOT$\}$ 18) (2018)

\bibitem{su2019one}
Su, J., Vargas, D.V., Sakurai, K.: One pixel attack for fooling deep neural
  networks. IEEE Transactions on Evolutionary Computation  \textbf{23}(5),
  828--841 (2019)

\bibitem{sun2019fine}
Sun, C., Qiu, X., Xu, Y., Huang, X.: How to fine-tune bert for text
  classification? In: China National Conference on Chinese Computational
  Linguistics. pp. 194--206. Springer (2019)

\bibitem{sundararajan2017axiomatic}
Sundararajan, M., Taly, A., Yan, Q.: Axiomatic attribution for deep networks.
  In: International conference on machine learning. pp. 3319--3328. PMLR (2017)

\bibitem{vaswani2017attention}
Vaswani, A., Shazeer, N., Parmar, N., Uszkoreit, J., Jones, L., Gomez, A.N.,
  Kaiser, {\L}., Polosukhin, I.: Attention is all you need. Advances in neural
  information processing systems  \textbf{30} (2017)

\bibitem{xie2017adversarial}
Xie, C., Wang, J., Zhang, Z., Zhou, Y., Xie, L., Yuille, A.: Adversarial
  examples for semantic segmentation and object detection. In: Proceedings of
  the IEEE International Conference on Computer Vision. pp. 1369--1378 (2017)

\bibitem{yang2019xlnet}
Yang, Z., Dai, Z., Yang, Y., Carbonell, J., Salakhutdinov, R.R., Le, Q.V.:
  Xlnet: Generalized autoregressive pretraining for language understanding. In:
  Advances in neural information processing systems. pp. 5754--5764 (2019)

\bibitem{zbontar2021barlow}
Zbontar, J., Jing, L., Misra, I., LeCun, Y., Deny, S.: Barlow twins:
  Self-supervised learning via redundancy reduction. In: International
  Conference on Machine Learning. pp. 12310--12320. PMLR (2021)

\bibitem{zhu2019freelb}
Zhu, C., Cheng, Y., Gan, Z., Sun, S., Goldstein, T., Liu, J.: Freelb: Enhanced
  adversarial training for natural language understanding. arXiv preprint
  arXiv:1909.11764  (2019)

\bibitem{zhu2015aligning}
Zhu, Y., Kiros, R., Zemel, R., Salakhutdinov, R., Urtasun, R., Torralba, A.,
  Fidler, S.: Aligning books and movies: Towards story-like visual explanations
  by watching movies and reading books. In: Proceedings of the IEEE
  international conference on computer vision. pp. 19--27 (2015)

\end{thebibliography}
\end{document}